# Deep Learning vs. Gradient Boosting: Benchmarking state-of-the-art machine learning algorithms for credit scoring

*Marc Schmitt* [a, b, 1]

[a] *Department of Computer Science, University of Oxford, UK*
[b] *Department of Computer & Information Sciences, University of Strathclyde, UK*

**Abstract**

Artificial intelligence (AI) and machine learning (ML) have become vital to remain competitive for financial services companies around the globe. The two models currently competing for the pole position in credit risk management are deep learning (DL) and gradient boosting machines (GBM). This paper benchmarked those two algorithms in the context of credit scoring using three distinct datasets with different features to account for the reality that model choice/power is often dependent on the underlying characteristics of the dataset. The experiment has shown that GBM tends to be more powerful than DL and has also the advantage of speed due to lower computational requirements. This makes GBM the winner and choice for credit scoring. However, it was also shown that the outperformance of GBM is not always guaranteed and ultimately the concrete problem scenario or dataset will determine the final model choice. Overall, based on this study both algorithms can be considered state-of-the-art for binary classification tasks on structured datasets, while GBM should be the go-to solution for most problem scenarios due to easier use, significantly faster training time, and superior accuracy.

# 1 Introduction

During the last decade, the world has witnessed several technological advances, which had a tremendous impact on financial services companies. Data-driven decision-making based on artificial intelligence (AI) and machine learning (ML) has become indispensable in today's global and ultra-competitive economy (Borges, Laurindo, Spínola, Gonçalves, & Mattos, 2021; Duan, Edwards, & Dwivedi, 2019; Dwivedi et al., 2021; M. Schmitt, 2020). Those and other developments triggered a change in the market structure for lending businesses (Claessens, Zhu, Frost, & Turner, 2018). This led to the increased importance of prediction accuracy for credit assessments instead of only relying on post-mortem protection in the form of LGD reduction through collateral. A pure focus on prediction accuracy has the advantage of cash-flow-based lending without collateral requirements from borrowers (Frost, Gambacorta, Huang, Shin, & Zbinden, 2019), which can foster financial inclusion and is especially important for consumers in developing countries and small to medium-sized corporations. Those clients have usually no collateral and had traditionally limited access to debt capital (Bazarbash, 2019).

---

[1] Corresponding author. E-mail address: marcschmitt@hotmail.de

Overall, it would be of significant value for peer-to-peer (P2P) lenders, FinTechs, as well as traditional banks if they could increase the accuracy of the applicant's default probability to assign a correct credit score during the application process. The better a financial institution/lender can predict the default probability/credit score of certain applicants the better they can shield themselves from potential costly credit losses. Besides, misclassification also results in missed revenue if a potentially good customer is wrongly assumed to be of high credit risk. Machine learning (Jordan & Mitchell, 2015) plays a fundamental role in achieving this goal as it counters the default problem at the origin – the decision of whether or not to take on a loan applicant.

Logistic regression has been the standard method for binary classification and is widely used in financial institutions to assign risk classes to applicants. It has served as the main benchmark for years due to its simple application and accurate enough forecasts (Kraus, 2014). However, machine learning has been proven to achieve superior performance against generalized linear models very early (Keramati & Yousefi, 2011) and several models such as decision trees, random forest, gradient boosting, support vector machines, and neural networks (NN) have gained popularity and are increasingly used in practice (Bazarbash, 2019). An analysis of the current literature reveals a clear trend showing that the models competing for the highest prediction accuracy are gradient boosting machines (GBM) and deep neural networks (DNN) – aka Deep Learning (DL).

Hamori et. al (2018) conducted an interesting empirical study analyzing the performance of ensemble learning in comparison to deep learning. The authors test different DL configurations by switching activation functions (ReLU, Tanh) and come after 100 test runs – which are averaged – to the conclusion that GBM does outperform DL. Those findings are in line with other papers (Addo, Guegan, & Hassani, 2018; Gunnarsson, vanden Broucke, Baesens, Óskarsdóttir, & Lemahieu, 2021), but there is also a body of literature that favors DL and comes to the conclusion that DL is superior to analytics use case in business (Kraus, Feuerriegel, & Oztekin, 2019; Lessmann, Baesens, Seow, & Thomas, 2015). Nevertheless, the literature is quite clear on the subject and we can conclude that there are currently only two models that compete w.r.t to predictive accuracy in credit scoring: Deep neural networks and gradient boosting machines.

Surprisingly, there is not a single study exclusively focusing on DL and GBM when it comes to credit scoring. It is usually a mix of several machine learning models and there is a clear gap in a direct comparison of GBM and DL for assessing the correct default category of a loan applicant. Also, most studies within the credit risk domain have relied on only one dataset to benchmark different algorithms w.r.t to prediction accuracy (Addo et al., 2018; Hamori et al., 2018). The most comprehensive study so far is from (Gunnarsson et al., 2021). Hence, it might be that the correct model choice is dependent on the underlying dataset.

The goal of this study is a direct comparison of GBM and DL in terms of prediction accuracy to correctly classify the default category of a customer. Three distinct datasets with different features have been used to account for the possibility that model choice/power is based on the characteristics of the underlying dataset. In doing so this study shed light on the predictive power and usefulness of both models for credit risk management within the lending market – specifically credit scoring.

The structure of this article is as follows. Section 2 "Theory and Methods" gives a quick overview of machine learning followed by an introduction of the methods used for the following empirical study: Gradient Boosting Machine (GBM) and Deep Learning (DL). Section 3 "Experimental Design" introduces the datasets and the process of model tuning (hyperparameter optimization). Section 4 "Numerical Results" presents the findings of this study as detailed as possible. Section 5 "Discussion of Results" analyses those findings and discusses their implications. Section 5 presents future research ideas, and the last section 6 ends with a conclusion.

## 2 Theory and methods

### 2.1 Machine Learning

The discipline of Machine Learning (Jordan & Mitchell, 2015) can be split into three major categories: Supervised learning; Unsupervised Learning; and Reinforcement Learning. The last one would be semi-supervised learning, which is essentially a hybrid of supervised and unsupervised ML. The most common method deployed in finance and most other industries is supervised learning (Ng, 2018). Supervised Learning requires labeled data to train a model, which will then be used to make predictions from new (unseen) data. Within the domain of supervised learning regression and classification can be distinguished. A classification problem - which is the method used in this study – has discrete outputs, where $Y \in \{0, ..., n\}$, which are categorical. The two models compared in this study within the context of default categorization are Gradient Boosting Machines and Deep Learning and are introduced next. See figure 1.

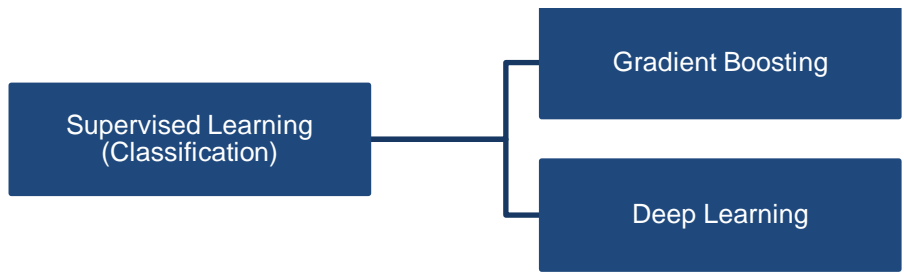

**Figure 1.** Given the contradictory literature regarding the prediction accuracy of gradient boosting and deep learning, the purpose of the experiment was a direct performance comparison of those two models within the context of credit scoring.

## 2.2 Deep Learning

Recent advances in AI research have improved the capabilities of Artificial Neural Networks and a new paradigm of Deep Learning was born (LeCun, Bengio, & Hinton, 2015). The three factors that helped DL to become mainstream are advances in data availability (Big Data), processing power (GPUs), and optimization algorithms (Goodfellow, Bengio, & Courville, 2016). One of the major advantages of DL is its ability to work with unstructured datasets, which improved many tasks and brought breakthroughs in text, speech, image, video, and audio processing (LeCun, et al., 2015). In 2014 the South Korean Go champion Lee Sedol was defeated by Deep Minds AlphaGo (Silver et al., 2016). This initial success of deep reinforcement learning was soon followed by AlphaGo Zero (Silver et al., 2017) and several other gaming-related multimedia appearances such as StarCraft (Pang et al., 2019) and Dota 2 (Katona et al., 2019). Deep learning did not only help AI to increase its popularity, but it also has a wider range of possible applications and is seen as one of the most disruptive technologies since the inception of the internet itself (Goodfellow, et al., 2016). Soon, the business world picked up on those developments and DL was increasingly used to enhance existing business analytics functions, including credit risk management (Bughin et al., 2017; Chui et al., 2018)

Deep Learning has many architectures. Examples are feed-forward artificial neural networks (ANN), Convolutional neural networks (CNNs), as well as Recurrent Neural Networks (RNNs). The architecture used in this empirical study is a multi-layer feedforward artificial neural network, which is in general the go-to algorithm structure for tabular data (Candel & LeDell, 2019). See figure 2 for a depiction of the architectural graph of a feed-forward neural network.

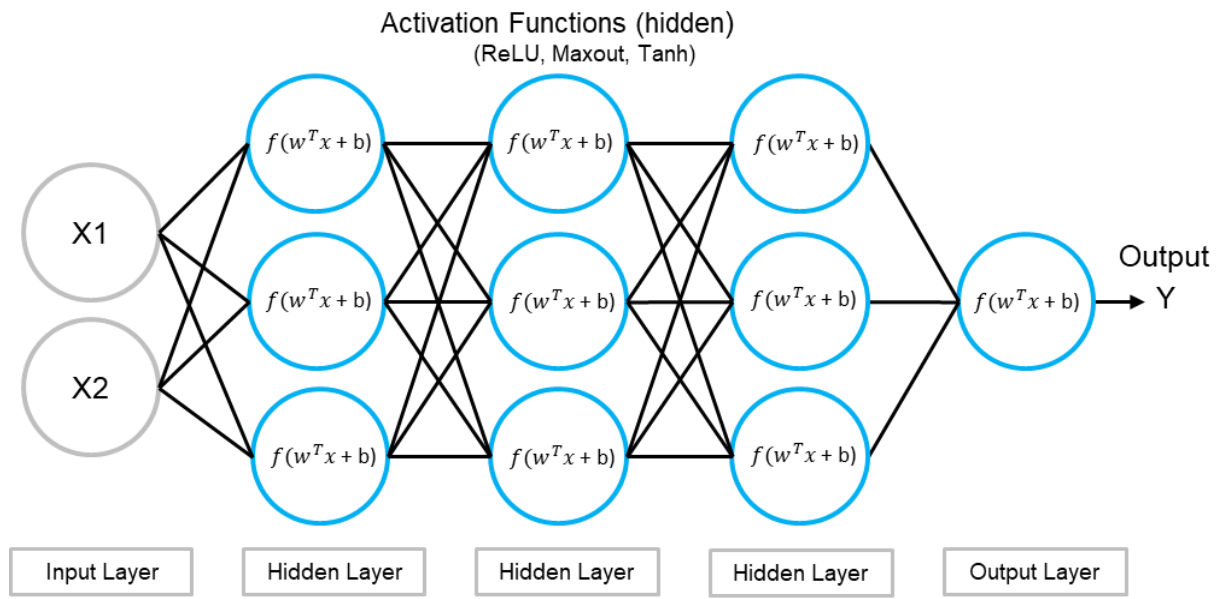

**Figure 2.** The feedforward artificial neural network is the DL algorithm of choice for this empirical study, which is based on structured datasets. It consists of one input as well as one output layer and various hidden layers. A linear combination of input variables and weights is fed into an activation function at each of the nodes to calculate a new set of values for the next layer. A neural network is trained with stochastic gradient descent and backpropagation (Goodfellow et al., 2016).

In a nutshell, a neural network multiplies the input features by a weight matrix and applies an activation function to calculate the input for the next layer. This procedure is repeated through the entire network. Specifically, the input variables $X_i = (X_1, X_2, X_3, ..., X_n)$ are fed into the neural network, weights $W_i = (W_1, W_2, W_3, ..., W_n)$ are added to each of those inputs, followed by the calculation of the linear combination $\sum X_i W_i = w^T x$. The resulting linear combination including a bias term is fed into an activation function to arrive at the output Y. This operation is repeated for every node in the network until the final activation function in the output layer has been reached. Three different activation functions for the hidden layers were utilized in this empirical study:

- the rectified linear unit (ReLU): $g(z) = \max(0, z) \in [0, \infty)$,

- the hyperbolic tangent function (Tanh): $g(z) = \frac{e^z - e^{-z}}{e^z + e^{-z}} \in [-1, 1]$, and the

- the Maxout function: $g(z) = \max(w^k z + b^k) \in (-\infty, \infty)$, $k \in \{1, ..., K\}$.

The activation function most widely used is the rectified linear unit (ReLU). As developments in DL are quite fast, I recommend checking the best/most common approaches w.r.t architectures as well as the concrete choice of the activation function to solve different problems regularly. Due to the research scope of this study – binary classification on structured data – the output activation function used is always the sigmoid function $sigm(z) = \frac{1}{1+e^{-z}} = \frac{e^z}{e^z+1} \in [0,1]$. A switch occurs only at the hidden layer.

## 2.3  Gradient Boosting

GBM is state-of-the-art when it comes to accuracy, especially for supervised learning problems on structured datasets (Ng, 2018). The first boosting algorithm – AdaBoost – was introduced by Freund and Schapire (1997). Four years later Friedman (2001) introduced the Gradient Boosting Machine, which is a more general form of the earlier algorithm due to the possibility to switch the loss function, which makes the AdaBoost algorithm essentially just a subset of the GBM introduced by Friedman (2001).

Boosting belongs together with bagging (Breiman, 1996) and stacking (Caruana, Niculescu-Mizil, Crew, & Ksikes, 2004) to the family of ensemble learning techniques and builds models in sequential order. The goal of ensemble learning is to combine multiple ML algorithms to achieve better predictive performance. The specific idea of boosting is to start with a so-called weak learner – a model only slightly better than random guessing – that gradually improves by correcting the error of the previous model at each step. See figure 3.

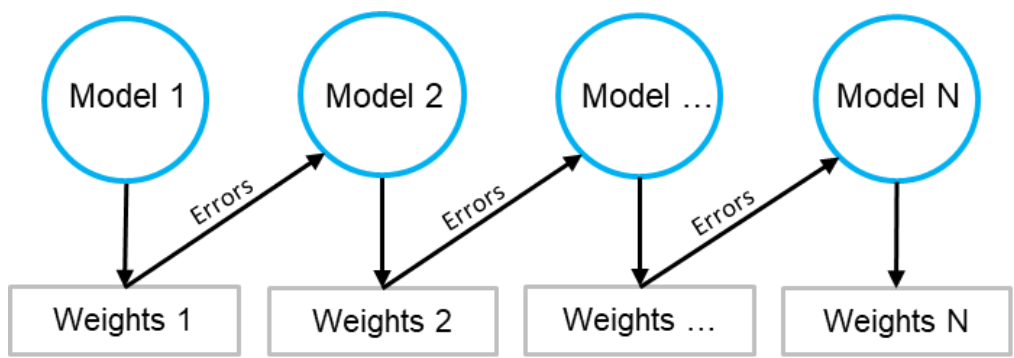

**Figure 3.** Gradient Boosting starts with a weak learner, typically a decision tree, and improves upon this initial learner iteratively at each step by correcting for the error of the previous model. GBM is one of the best performing ML models currently available.

The most common form of boosting uses decision trees and sequentially ads one tree at a time. This step-by-step adjustment forces the model to gradually improve its performance and leads to higher accuracy (Hastie, Tibshirani, & Friedman, 2017). There are several different gradient boosting implementations out there. This study uses the gradient boosting version implemented by Malohlava and Candel (2019) which is based on Hastie et al. (2017).

## 2.4 AUC as Evaluation Measure

To determine the predictive power of an ML model, an evaluation method is required. There are several valuation measures available in ML. In this experiment, the area under the receiver operating characteristics (ROC) curve (AUC) is used to measure the accuracy of the models. ROC is based on a concept called confusion matrix which is a contingency table and is frequently used in binary classification problems. See figure 4.

|  |  | Prediction | |
|---|---|---|---|
|  |  | Good | Bad |
| Actual | Good | True Positive (TP) | False Negative (FN) |
| Actual | Bad | False Positive (FP) | True Negative (TN) |

**Figure 4.** A confusion matrix is a basic ingredient for the ROC Curve. It shows the connection between true positives and negatives and false positives and negatives.

True positives (TP) represent good observations that were classified as such, whereas false positives (FP) are observations that were incorrectly classified as good. The same logic applies to true negatives (TN), which represent the predicted values correctly classified as bad, whereas the false negatives (FN) represent observations incorrectly classified as bad.

Precision is defined as the ratio of true positives over the sum of true positives and false positives.

$$Precision\ (FPR) = \frac{TP}{TP + FP} \qquad (1)$$

Recall is defined as the ratio of true positives over the sum of true positives and false negatives.

$$Recall\ (TPR) = \frac{TP}{TP + FN} \qquad (2)$$

As can be seen in figure 5, the FPR represents the x-axis and the TPR represents the y-axis of the ROC plot. A perfect model would have a TPR of 1 and an FPR of 0.

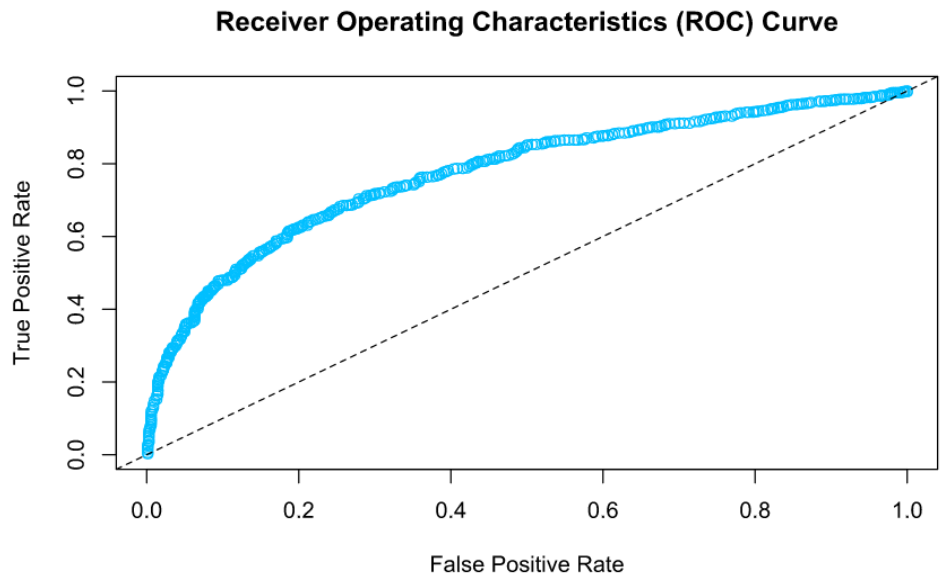

**Figure 5.** The AUC of the ROC Curve is an accuracy measure for classification problems and will help to assess the predictive power of the following models.

The closer the AUC comes to 1 the stronger and more accurate is the model. The AUC as a comparison measure is only valid when the underlying distribution is uniform, which means the outcome of each class is equally likely (Flach, Hernández-Orallo, & Ferri, 2011). The models first predict the probability of default for every observation (customer) and based on a decision boundary assigns a classification to the customer of either good (1) or bad (0). The default decision boundary for binary classification problems as in this case is 50% and is represented by a diagonal line from (0,0) to (1,1) of the ROC plot. Hence if we received an AUC of 50% or below our model would be purely random and therefore worthless.

## 3  Experimental Design

Financial institutions use credit risk models as scoring models to determine a client's default probability. These estimates help to decide whether a certain customer should receive a loan or a credit card. The primary objective of this study is to benchmark the above introduced predictive models (Gradient Boosting and Deep Learning) to correctly classify the default category of a customer. Due to the difficulty of determining the best activation function choice three different DL configurations with different activation functions (ReLU, Maxout, Tanh) are run. Only the activation functions in the hidden layers are switched. Due to the research scope - binary classification on structured datasets - the loss function (binary cross-entropy) and output activation function (sigmoid) remain the same. Three distinct datasets containing detailed client data are used.

## 3.1 Data and Preprocessing

The rationale for the used datasets is their relevance to credit risk management. One part of credit risk management is concerned with assessing the likelihood that a counterparty (e.g. loan applicant) will not be able to repay its obligation in part or in full. For this purpose, the output of credit risk models is either the probability of default or a credit score, which can be binary or multi-class depending on the specific use case. Credit scoring in this paper refers to the binary classification of loan applicants, and either a "good" or "bad" label is assigned to the counterparty.

The datasets are from existing retail banks and have been widely used in earlier studies (Gunnarsson et al., 2021; Guo, He, & Huang, 2019; Hamori et al., 2018; Lessmann et al., 2015; Teng, He, Xiao, & Jiang, 2013; Yeh & Lien, 2009). The contained features resemble the typical information available to a retail bank and are therefore valid as real-world examples. See table 2 for a detailed description of the features contained in datasets 1 and 2.

The 3 datasets resemble real-world customer data, are all publicly available and can be downloaded from the UCI Machine Learning Repositories, which makes the reproducibility of this empirical analysis possible. The datasets contain 23, 20, and 14 features respectively, which are historical client data and will serve as predictor variables to calculate the default category of each observation. All 3 datasets contain a target column that identifies whether or not the client defaulted. See table 1 for the details of each dataset including the resampling information.

**Table 1.** Description of Datasets

| Dataset | Observations | Good | Bad | Balanced (Experimental Setup) | Features |
|---|---|---|---|---|---|
| Taiwan | 30,000 | 23,364 | 6,636 | 6,636/6,636 | 23 |
| Germany | 1,000 | 700 | 300 | 300/300 | 20 |
| Australia | 690 | 307 | 383 | 307/307 | 14 |

The empirical study is based on 3 data sets, each containing several features (predictors) including a target column containing the default information (response). The datasets have been resampled and balanced (equal ratio of good and bad clients) to avoid the tendency for the AUC metric to favor the majority class.

**Dataset 1 – Taiwan:** The first dataset represents payment information from Taiwanese credit card clients. It was first used by Yeh and Lien (2009) and contains 30,000 observations where 6,636 are flagged as defaults. The dataset contains mainly historical payment information.

Each observation (or feature set) contains 23 features including a binary response column for the default information of the credit cardholder.

**Dataset 2 – Germany:** The second dataset represents detailed customer-level data from a German bank and contains 1,000 observations where 300 are flagged as defaults. Each observation contains 20 features across a diver's range of categories including a binary response column that indicates whether or not a particular client defaulted on their loan payments.

**Dataset 3 – Australia:** The third dataset contains data for credit card applications for clients based in Australia. The dataset contains 690 observations where 383 are flagged as bad. Each observation contains 14 features including a binary response column indicating whether or not the person defaulted. The attribute names and values in this dataset have been changed to meaningless symbols due to confidentiality reasons.

See table 2 below for a more detailed description of the specific features contained in datasets 1 and 2. No such description exists for the third dataset due to confidentiality reasons.

**Table 2.** A detailed description of features contained in dataset 1 and 2

| | Dataset 1 - Taiwan | | Dataset 2 - German |
|---|---|---|---|
| **Variable** | **Description** | **Variable** | **Description** |
| X1 | Amount of the given credit | X1 | Balance of checking account |
| X2 | Gender (1 = male; 2 = female) | X2 | Duration in months |
| X3 | Education* | X3 | Credit history |
| X4 | Marital status** | X4 | For what was the loan taken |
| X5 | Age (year) | X5 | Credit amount |
| X6 | Payment history September 2005 | X6 | Savings account plus bonds |
| X7 | Payment history August 2005 | X7 | Duration of current employment |
| X8 | Payment history July 2005 | X8 | Installment rate as % of income |
| X9 | Payment history June 2005 | X9 | Marital status and gender |
| X10 | Payment history May 2005 | X10 | Other debtors/guarantors |
| X11 | Payment history April 2005 | X11 | Present residence since |
| X12 | Amount of bill statement in Sep 2005 | X12 | Type of owned properties |
| X13 | Amount of bill statement in Aug 2005 | X13 | Age of applicant |
| X14 | Amount of bill statement in Jul 2005 | X14 | Housing (rent, own, free) |
| X15 | Amount of bill statement in Jun 2005 | X15 | Credits at other banks |
| X16 | Amount of bill statement in May 2005 | X16 | Existing credits at this bank |
| X17 | Amount of bill statement in Apr 2005 | X17 | Employment/Level of qualification |
| X18 | Amount paid September 2005 | X18 | The number of dependents |
| X19 | Amount paid August 2005 | X19 | Registered telephone or none |
| X20 | Amount paid July 2005 | X20 | Immigrant/foreign worker |
| X21 | Amount paid June 2005 | | |
| X22 | Amount paid May 2005 | | |
| X23 | Amount paid April 2005 | | |

* (1 = graduate school; 2 = university; 3 = high school; 4 = others)

** (1 = married; 2 = single; 3 = others)

Several adjustments were undertaken during the preprocessing phase to better serve the purpose of this study:

Random under-sampling was used to create a balanced data set for this classification study. Imbalanced datasets can result in a bias towards the majority class. The accuracy measure used in this paper – Area under the curve (AUC) – is more reliable when the model is trained with a balanced dataset. Predictive models that strive for maximum AUC tend to gravitate towards a classification that overrepresents the majority category which results automatically in higher prediction accuracy. If the dataset would have an imbalance of 80 to 20, the ML algorithm could always achieve an 80% accuracy without having true predictive power. To address this problem the class distribution (the ratio between the two categories good and bad) was adjusted and brought to a state of equilibrium. This is done by under-sampling the majority class and by doing so creating an equal ratio of good and bad observations (categories). The subset of good observations was randomly drawn from the total population.

Another important step during the preprocessing of the data is to replace categorical values with a numerical representation. For example, dataset 1 contains only three numerical values which can directly be used to fit the machine learning models. The other features are categorical and had to be transformed into factor variables to be processed. This is often done by a method called one-hot encoding. One-hot encoding is widely used for classifying categorical data and transforms categorical labels into vectors of zeros and ones. The length of the resulting vectors is equal to the number of categories where each element within the vectors corresponds to one of those categories. This method potentially results in a significant increase in the feature set depending on the number of categories as well as the number of elements within each category. This was done for dataset 2 and 3. Dataset 1 is already clean and consists of only numerical values. Significant preprocessing is not required. One last step – that was done for all datasets – was to change the response variable from a numeric representation to a binary factor which is necessary for a classification problem.

This study uses a training set, a validation set, and a test set. The first scenario which uses a training set size of 80% (80:10:10) is further referred to as 80:20 split while the second scenario using a training set size of 70% (70:15:15) is referred to as 70:30 split.

### 3.2 Hyperparameter Settings

Machine learning is an empirical process that involves trial, error, and experimentation. Hyperparameter optimization or model tuning describes the process of finding the optimal combination of hyperparameters for a machine learning algorithm. It is a multidimensional optimization problem and becomes more computationally demanding with an increasing number of parameters.

All models were carefully tuned to reach a performance that is adequate for the comparison in this study. In the case of DL, three different models were trained each containing a different activation function in the hidden layers (ReLU, Tanh, and Maxout) while holding all other parameters constant. The H2O framework allows the user to choose the activation functions (ReLU, Maxout, Tanh) for the hidden layers, the appropriate loss function, and the response type.

The H2O deep learning framework uses the above-specified activation function throughout the network (hidden layers) and based on the response column (binary) and loss function choice

(cross-entropy) determines the appropriate activation function (in this case sigmoid) for the final layer.

Dropout has been shown to improve accuracy (Srivastava, Hinton, Krizhevsky, Sutskever, & Salakhutdinov, 2014); hence all 3 DL models use a dropout ratio of 0.50. The concrete hyperparameter settings for each dataset and model in the case of the 80:20 split can be found in table 3.

**Table 3.** Hyperparameter setting of GBM and DL for the 80:20 split

| Dataset | GBM - Parameters | Value | DL - Parameters | Value |
|---|---|---|---|---|
| Taiwan | ntrees | 30 | activation* | ReLU, Tanh, Maxout |
|  | max_depth | 5 | hidden | c(200, 200, 200) |
|  | min_rows | 10 | epochs | 50 |
|  | learn_rate | 0,2 | rate | 0,2 |
| Germany | ntrees | 400 | activation* | ReLU, Tanh, Maxout |
|  | max_depth | 30 | hidden | c(200, 200) |
|  | min_rows | 2 | epochs | 15 |
|  | learn_rate | 0,05 | rate | 0,01 |
| Australia | ntrees | 10 | activation* | ReLU, Tanh, Maxout |
|  | max_depth | 15 | hidden | c(200, 200) |
|  | min_rows | 10 | epochs | 15 |
|  | learn_rate | 0,01 | rate | 0,003 |

*These are the activation functions for the hidden layers

The hyperparameter settings of GBM and DL in the case of the 70:30 split can be found in table 4.

**Table 4.** Hyperparameter setting of GBM and DL for the 70:30 split

| Dataset | GBM - Parameters | Value | DL - Parameters | Value |
|---|---|---|---|---|
| Taiwan | ntrees | 30 | activation* | ReLU, Tanh, Maxout |
|  | max_depth | 5 | hidden | c(100, 100) |
|  | min_rows | 10 | epochs | 12 |
|  | learn_rate | 0,2 | rate | 0,2 |
| Germany | ntrees | 390 | activation* | ReLU, Tanh, Maxout |
|  | max_depth | 24 | hidden | c(200, 200) |
|  | min_rows | 2 | epochs | 15 |
|  | learn_rate | 0,095 | rate | 0,01 |
| Australia | ntrees | 11 | activation* | ReLU, Tanh, Maxout |
|  | max_depth | 13 | hidden | c(200, 200) |
|  | min_rows | 7 | epochs | 15 |
|  | learn_rate | 0,01 | rate | 0,003 |

*These are the activation functions for the hidden layers

Searching the complete parameter space is computationally demanding. According to Bergstra and Bengio (2012) and Ng (2018), random search results in the best parameter choices compared to grid search and can be completed faster. Hence, an exhaustive grid search is not adequate considering the tradeoff between accuracy and training time. To tune the hyperparameters in this study a combination of random search, grid search, and manual adjustments was used. The random grid search was applied for 15 minutes. Afterward, a grid search was used around a small interval of the parameters determined by random search to further calibrate the models and improve accuracy. Once this was completed, I tried micro-adjustments of the hyperparameters to see whether there is a possibility left to enhance the accuracy levels. The manual changes were only done for one parameter at a time while holding all the others constant. Where this was possible it did only impact the accuracy levels slightly. No significant performance improvement could be reached at the final step and mainly the random search plus selective grid search resulted in maximum performance.

In case the hyperparameter value is not mentioned in Table 3 or 4 the default value ascribed by H2O was used during the model training. Concrete advice on parameter choices in ML is subject to further research (Ng, 2018).

### 3.3 Software

The complete software setup – data preparation, preprocessing, model fitting, and evaluation – was done in the integrated development environment (IDE) RStudio, which is widely used for research in data science and machine learning and based on the statistical programming language R (R Core Team, 2019). The major package utilized in this study to develop the machine learning models is H2O, which is a powerful open-source ML platform. It is written in Java and offers a multitude of predictive models (LeDell & Gill, 2019). One of the core advantages is the high abstraction level to focus on the experiment itself. R is connected to H2O through a REST API (Aiello, et al., 2016).

## 4 Numerical Results

Three datasets containing detailed customer-level data were used to benchmark Gradient Boosting Machine (GBM) against Deep Learning (DL). Table 5 shows the out-of-sample performance of the trained GBM and DL models and gives a complete summary of the results obtained during this study. It shows the AUC for each of the 3 datasets as well as the two training/test set splits. The AUC as accuracy measures should be diagnostically conclusive as the datasets were resampled and balanced before the model training.

**Table 5.** Model results separated by dataset as well as training/test set split

| Dataset | Method | AUC | Method | AUC |
|---|---|---|---|---|
| | 80:20 Split | | 70:30 Split | |
| Taiwan | Gradient Boosting | 0.773 | Gradient Boosting | 0.771 |
| | DL + ReLu | 0.765 | DL + ReLu | 0.759 |
| | DL + Tanh | 0.744 | DL + Tanh | 0.741 |
| | DL + Maxout | 0.761 | DL + Maxout | 0.754 |
| Germany | Gradient Boosting | 0.885 | Gradient Boosting | 0.823 |
| | DL + ReLu | 0.941 | DL + ReLu | 0.838 |
| | DL + Tanh | 0.919 | DL + Tanh | 0.816 |
| | DL + Maxout | 0.936 | DL + Maxout | 0.829 |
| Australia | Gradient Boosting | 0.988 | Gradient Boosting | 0.989 |
| | DL + ReLu | 0.964 | DL + ReLu | 0.961 |
| | DL + Tanh | 0.961 | DL + Tanh | 0.943 |
| | DL + Maxout | 0.974 | DL + Maxout | 0.965 |

## 4.1 Dataset 1: Taiwan

In the case of the 80:20 split GBM achieved an AUC of 0.773 during the out-of-sample test. The best performing DL model used the ReLU activation function and achieved an AUC of 0.765. See figure 6.

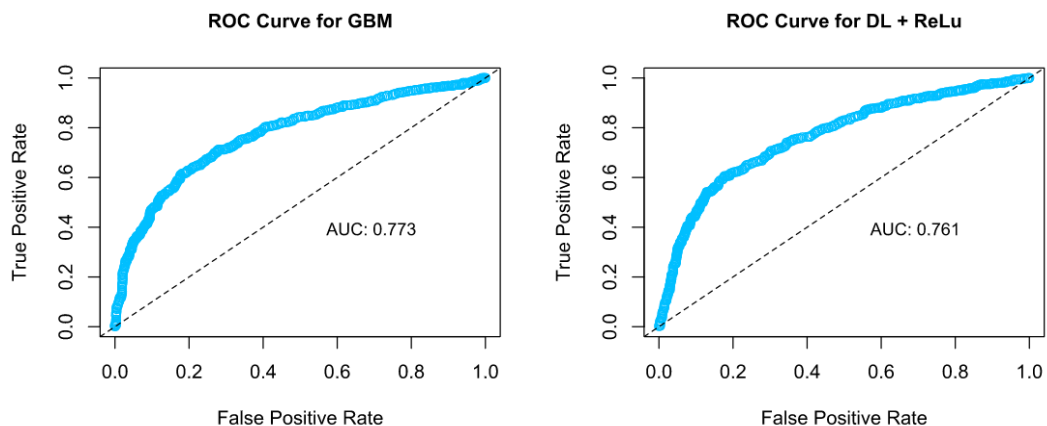

**Figure 6.** Performance of GBM vs. DL with ReLU function on Taiwanese dataset and 80:20 split

The 70:30 split has a slightly lower classification power in terms of the used metric, which shows an AUC of 0.771 for GBM and an AUC of 0.759 for the DL model, which also used the ReLU as the activation function. See figure 7.

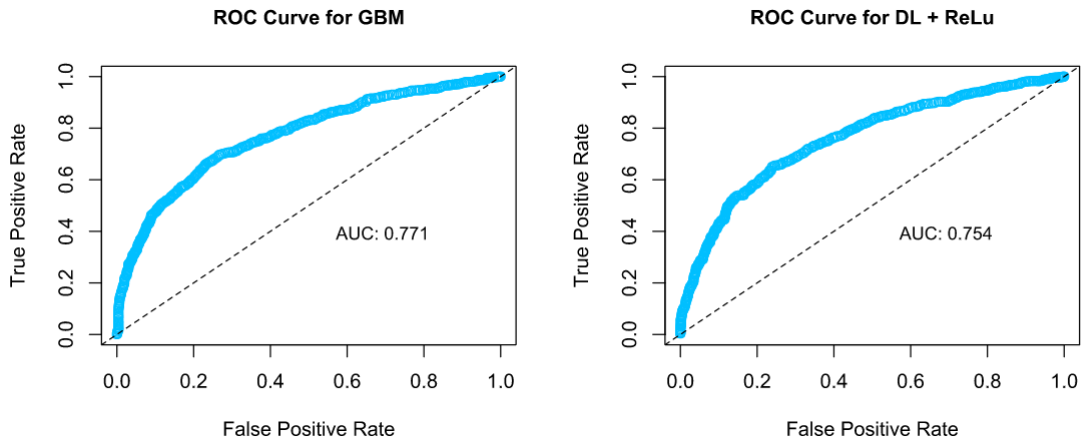

**Figure 7.** Performance of GBM vs. DL with ReLU function on Taiwanese dataset and 70:30 split

Overall, based on the first dataset GBM remains in terms of accuracy the superior model compared to DL. The DL model utilizing the Tanh activation function was at the lowest performance end at both training/test set splits.

## 4.2 Dataset 2: Germany

The results for Dataset 2 are surprisingly different from Dataset 1. In the case of the 80:20 split GBM achieved an AUC of 0.885 on the test data, which was significantly lower than the best performing DL which achieved an AUC of 0.941. See figure 8.

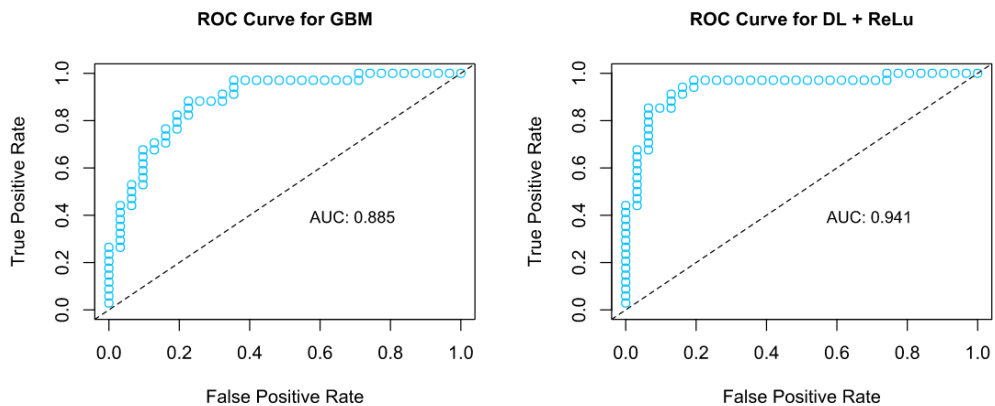

**Figure 8.** Performance of GBM vs. DL with ReLU function on German dataset and 80:20 split

The 70:30 split for dataset 2 has a slightly different classification accuracy with an AUC of 0.823 for GBM and an AUC of 0.838 for the best DL model that uses again the ReLU activation function. See figure 9.

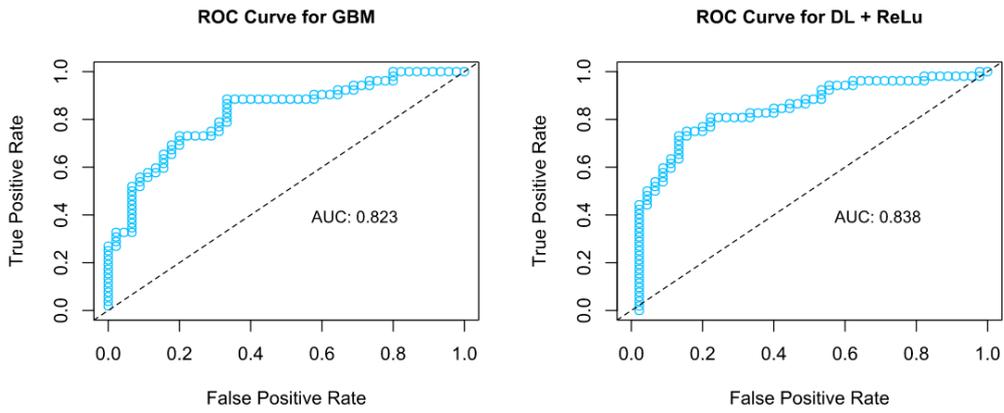

**Figure 9.** Performance of GBM vs. DL with ReLU function on German dataset and 70:30 split

Overall, the position for the highest classification power is reversed in the case of the second dataset and DL was able to outperform GBM in terms of the accuracy metric AUC. In addition, all 3 DL models could represent the dataset better than GBM, while the DL model with the Tanh activation function takes the lowest spot w.r.t to classification accuracy.

### 4.3  Dataset 3: Australia

The out-of-the-sample results obtained for Dataset 3 in the case of the 80:20 split show an AUC of 0.988 for GBM, and an AUC of 0.974 for the best DL model, which was this time achieved by the Maxout activation function. See figure 10.

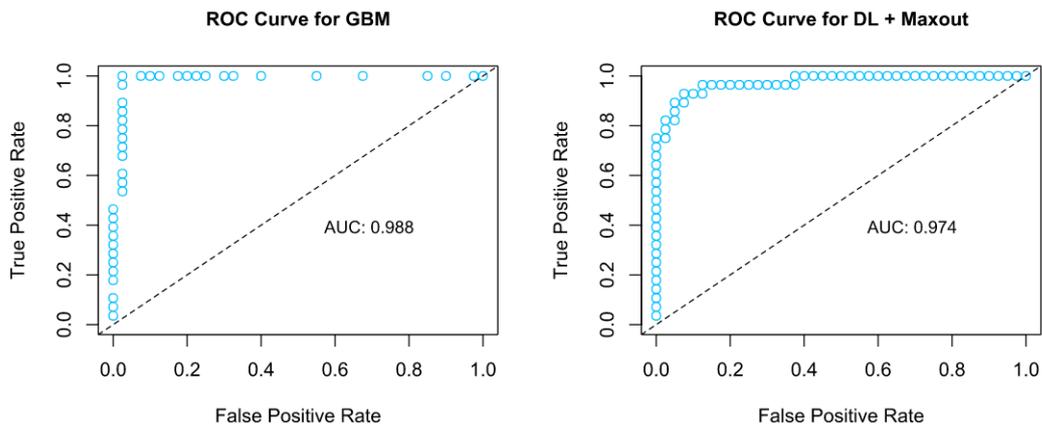

**Figure 10.** Performance of GBM vs. DL with Maxout function on Australian dataset and 80:20 split

In the case of the 70:30 split GBM achieved a similarly impressive AUC of 0.989 on the test data, while the best performing DL model which also uses the Maxout activation function achieved an AUC of 0.965. See figure 11.

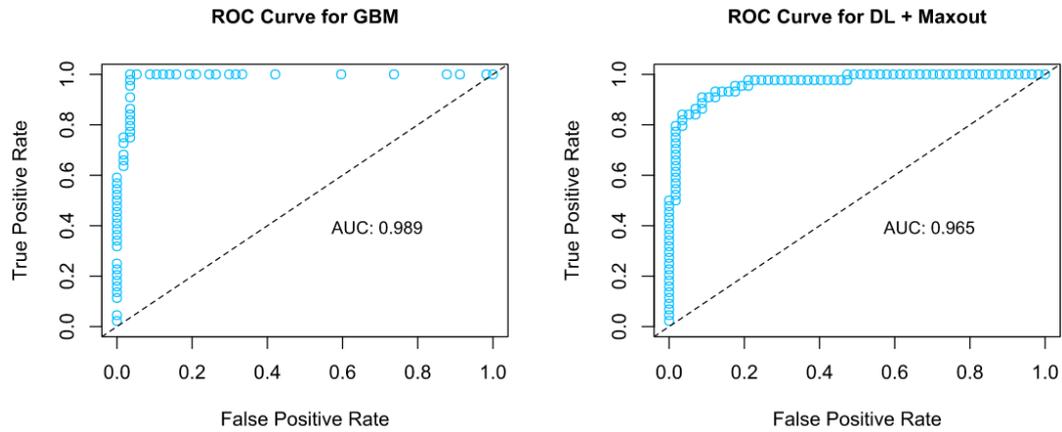

**Figure 11.** Performance of GBM vs. DL with Maxout function on Australian dataset and 70:30 split

Overall, the performance of GBM is superior to DL in terms of the AUC performance metric for both training/test set splits. Again, the DL model with the Tanh activation function had the lowest AUC among the DL models in both cases.

## 5  Discussion of Results

The results of this study clearly show that GBM is superior in terms of prediction accuracy compared to DL. This is especially true in the case of dataset 1 (Taiwan) and dataset 3 (Australia). Those findings confirm the conclusions reached by earlier studies (Addo et al., 2018; Gunnarsson et al., 2021; Hamori et al., 2018). Nevertheless, DL was able to outperform GBM on the second dataset, which suggests that the underlying structure of datasets is important and only slight variations might result in the need for a different model or at least a change in the model configuration.

Given the observation that there exists a tendency to achieve higher prediction accuracy when increasing the number of observations within the training set, it is recommended to use as many observations in the training set as possible. However, the size of the two datasets Germany and Australia with only 1000 and 690 observations respectively are relatively small. Especially in combination with the additional resampling and balancing, which further reduced the number of observations to 600 and 614 made it difficult to go for a 90:10 split. The resulting number of observations in the test set would have been too small, which would have resulted in inadequate performance.

The choice of the activation function for deep learning does indeed have an impact on model performance. Comparing the different DL models and activation function choices, it can be observed that the ReLU activation function performs the best, with the Maxout activation function as a close follower. The Tanh activation function has shown a consistent underperformance across all the tested scenarios.

The impact of the different training/test set splits (80:20 and 70:30) on model performance was quite trivial and did not alter the essential findings across all three datasets. Overall, the training process GBM is less complex, and it is way easier to achieve a satisfactory prediction accuracy with Gradient Boosting Machine than with Deep Learning. Also, the space of hyperparameters for GBM is smaller and fewer variations have to be considered, which results in significantly faster training time for GBM compared to DL.

Decisions have often to be done in real-time, which requires model adjustments to be carried out within minutes or maximum hours. If two or more models deliver more or less the same result the less complex model should be the preferred choice. This becomes particularly important for large datasets. Flexibility and fast reaction times are key in many business functions. Feature engineering and hyperparameter tuning over several days or weeks to arrive at a satisfactory result is not realistic, which is another reason why GBM should be favored over DL.

There is an overall consensus that GBM is superior to DL when it comes to structured (tabular) data sets and DL is dominating tasks based on unstructured datasets (Ng, 2018). The findings of this study are in line with the current literature but do not suggest a complete switch to GBM. However, papers that propose deep learning as the superior solution seem not to represent reality. Machine learning remains an empirical process. It is therefore recommended to test different models for different datasets to find the one model that can best represent the information contained within the data. It is also advisable to reevaluate parameter settings and model choice after a non-trivial change in the fundamental dataset has occurred as this might result in different requirements w.r.t. model configurations.

Overall, based on this study both algorithms can be considered state-of-the-art for binary classification tasks on structured datasets, while GBM should be the go-to solution for most problem scenarios due to easier use, significantly faster training time, and superior accuracy.

To strengthen the above findings additional datasets can be used, but it is unlikely that the findings will be negatively challenged as results already indicate that there is no guarantee, but a strong tendency that GBM is the preferred choice for structured datasets in the case of binary classification. However, further research could successfully strengthen those findings and confirm that GBM is the best model available for structured datasets.

# 6   Future Research

The traditional logistic regression does not reach accuracy levels that come close to artificial neural networks or ensembles such as bagging and boosting. Despite this fact, those algorithms belong still the favorite tools in business analytics departments. Deep learning and gradient boosting are essentially black box models that are not explainable and hence difficult to communicate. Credit risk management within financial services is a field where model interpretability and the need for causal explanations are often dominating pure predictive ability. Further research should therefore focus on transparency, auditability, and explainability of machine learning models in credit scoring (Bücker, Szepannek, Gosiewska, & Biecek, 2022; Dastile & Celik, 2021).

The question of why DL was not able to find its way into business analytics functions as expected due to the hype and attention it has received recently is another interesting question (M. A. Schmitt, 2022). The usual explanations are missing skills due to the war of talent and regulatory requirements – especially in finance and insurance. However, there might be several other explanations for why the adoption is lacking behind expectations (Kar, Kar, & Gupta, 2021). The adoption speed of AI/ML models is especially relevant for incumbent corporations (Chui et al., 2018).

The primary goal of this empirical study was to analyze prediction accuracy for binary classification on structured data. However, credit scoring could largely move away from using traditional data for the assessment of the default probability. Not only is it possible to utilize a vast array of new data sources that grow consistently in volume, but our ability to harvest and store those data for improved decision-making has dramatically improved as well. Many FinTech companies have already started to use unstructured data (e.g., text mining and social media data) to further enhance the ability to assess the default classification of customers, which could further restructure the lending market. Another research direction could focus on the special characteristics of DL to enhance credit risk models with unstructured data to further increase the accuracy of lending decisions. This seems especially relevant due to significant market changes in developing countries that drive forward non-traditional borrowing as peer-to-peer lending (Wang, Han, Liu, & Luo, 2019).

# 7   Conclusion

The global economy has changed significantly over the last years and new entrants in the form of FinTech companies are increasingly disrupting the current lending market structure.

Advanced analytics – especially AI/ML models – are essential in this increasingly competitive market. GBM and DL are the two dominating forces currently shaping the credit risk landscape when it comes to supporting lending decisions. This study has shown that – in the case of structured datasets – GBM tends to be superior in terms of prediction accuracy. It is easier to use and has also the advantage of computational speed. However, DL was able to beat GBM in one case, which shows that the outperformance of GBM is not always guaranteed. It seems the model choice is also dependent on the concrete problem scenario and underlying characteristics of the dataset, and it might be wise to choose a predictive model that is best suited for the problem scenario at hand. Overall, DL and GBM are both powerful models to support financial services companies to calculate the default risk of a counterparty.